\pdfoutput=1

\documentclass[11pt]{article}

\usepackage[]{acl}

\usepackage{times}
\usepackage{latexsym}
\usepackage{textcomp}
\usepackage{float}
\usepackage{multirow}
\usepackage{booktabs}
\usepackage{graphicx}
\usepackage[T1]{fontenc}
\usepackage{courier}
\usepackage{mathtools}
\usepackage{url}

\usepackage[utf8]{inputenc}
\usepackage{subfig}
\usepackage[belowskip=-5pt,aboveskip=5pt]{caption}

\usepackage{microtype}
\usepackage{inconsolata}
\usepackage[figuresright]{rotating}

\usepackage{tgcursor}
\usepackage{xcolor}
\usepackage{xspace}
\usepackage[utf8]{inputenc}
\usepackage{authblk}
\usepackage{natbib}
\usepackage{lipsum}
\usepackage{booktabs, 
            multirow, threeparttable}
\usepackage{enumitem}
\usepackage{rotating}
\providecommand{\lb}{\mbox{$\langle$}}
\providecommand{\rb}{\mbox{$\rangle$}}
\usepackage{colortbl}

\usepackage[most]{tcolorbox}
\definecolor{ddgrey}{HTML}{D6D2CE}

\newtheorem{definition}{Definition}
\newtheorem{task}{Task}

\floatstyle{ruled}
\newfloat{listing}{tb}{lst}{}
\floatname{listing}{Listing}

\usepackage{xspace}
\newcommand{\method}{\texttt{TriSum}\xspace}

\title{\method: Learning Summarization Ability from Large Language Models with Structured Rationale}

\author[1]{\textbf{Pengcheng Jiang}}
\author[2]{\textbf{Cao Xiao}}
\author[1]{\textbf{Zifeng Wang}}
\author[2]{\textbf{Parminder Bhatia}}
\author[1]{\\\textbf{Jimeng Sun}}
\author[1]{\textbf{Jiawei Han}}
\affil[1]{University of Illinois at Urbana-Champaign}
\affil[2]{GE HealthCare}
\affil[1]{\texttt{\{pj20, zifengw2, jimeng, hanj\}@illinois.edu}}
\affil[2]{\texttt{danicaxiao@gmail.com}}

\begin{document}
\maketitle
\begin{abstract}
The advent of large language models (LLMs) has significantly advanced natural language processing tasks like text summarization. However, their large size and computational demands, coupled with privacy concerns in data transmission, limit their use in resource-constrained and privacy-centric settings. To overcome this, we introduce \method, a framework for distilling LLMs' text summarization abilities into a compact, local model. Initially, LLMs extract a set of aspect-triple rationales and summaries, which are refined using a dual-scoring method for quality. Next, a smaller local model is trained with these tasks, employing a curriculum learning strategy that evolves from simple to complex tasks. Our method enhances local model performance on various benchmarks (CNN/DailyMail, XSum, and ClinicalTrial), outperforming baselines by 4.5\%, 8.5\%, and 7.4\%, respectively. It also improves interpretability by providing insights into the summarization rationale.
\end{abstract}

\section{Introduction}
Large language models (LLMs), such as GPT-3 \citep{brown2020language} and its successors \citep{chowdhery2022palm, touvron2023llama, openai2023gpt4}, has greatly advanced natural language processing tasks, including machine translation \citep{brants2007large}, question-answering (QA) systems \citep{yang-etal-2019-end-end,bao-etal-2021-plato}, and text summarization \citep{liu2019text}. However, due to their substantial model size and computational demands, their utility can be limited in resource-constrained environments~\citep{strubell-etal-2019-energy}. 
Moreover, privacy becomes a major concern when sending proprietary data to external LLM services like ChatGPT.

\begin{figure}[t]

\centering
\includegraphics[width=\linewidth]{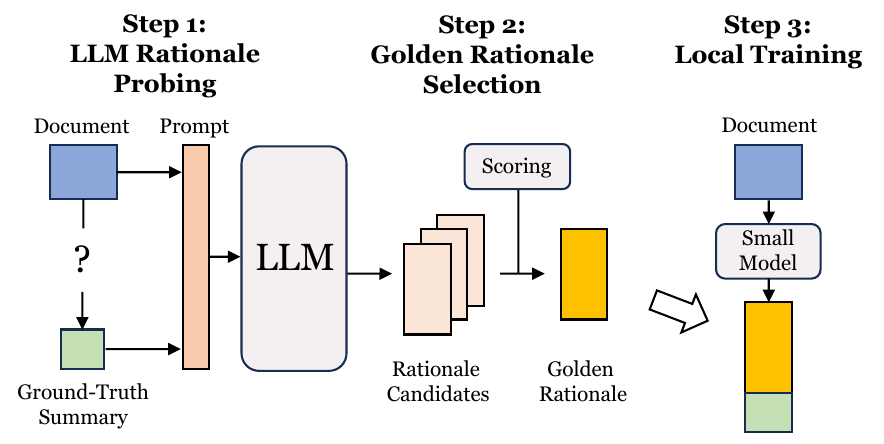}

\caption{A conceptual demonstration of our three-step framework \method that endows local small models with LLM's text summarization capability.}
\label{fig:intro}
\vspace{-1em}
\end{figure}

Among others, text summarization is a crucial task for transforming lengthy texts into concise yet informative summaries \citep{radev-etal-2002-introduction}. However, many existing methods struggle to generate structured summaries \citep{brown2020language, gekhman2023trueteacher, liu2023learning}. These structured summaries need to encompass essential aspects, key entities and relationships, and a coherent final summary derived from these aspects and rationales. Recent developments have seen the utilization of LLMs to grasp a text's topic structure and core ideas \citep{NIPS2017_3f5ee243, wei2023chainofthought}, suggesting their potential in generating structured text summaries. While rational distillation from LLMs has been employed for NLP tasks like QA, natural language understanding (NLU), and arithmetic reasoning \citep{wang2022pinto, hsieh2023distilling, magister2023teaching, ho2023large}, its applicability to abstractive text summarization remains unexplored.

In this study, we aim to distill LLMs' text summarization prowess into a more compact local model. We enhance the transparency and interpretability of this local model by incorporating elicited rationales from LLMs' summarization process as additional guidance. To achieve this, we introduce a three-step framework \method (as shown in Figure \ref{fig:intro}) involving LLM rationale probing, golden rationale selection, and local training:

\noindent \textbf{Step 1}: We first prompt vital \textit{aspect-triple} rationales and summaries from the input text using LLMs. This set includes essential aspects, relevant triples extracted from the text, and a concise summary that's tied to these aspects and triples. 

\noindent \textbf{Step 2}: Next, to ensure quality, we employ a dual-scoring method for selecting golden (high-quality) rationales to use in the subsequent training. This method evaluates the summary's quality based on semantic similarity and ensures coherent rationales using a topic distribution-based approach.

\noindent \textbf{Step 3}: Last, we train our compact local model using a curriculum learning approach \citep{nagatsuka-etal-2021-pre, xu-etal-2020-curriculum}. This method progressively fine-tunes the model by starting with simpler tasks and gradually advancing to more complex ones. This process enables our model to gradually incorporate the rationalized summarization skills acquired from the LLMs.

Our research brings the following contributions.

\begin{itemize}[leftmargin=*]
\item We introduce a new approach that distills LLMs' abstractive text summarization power into a small local model.
\item We design a scoring mechanism to select high-quality rationales, which serves as a robust base for training the local model.
\item Through extensive experiments we show that incorporating LLM-generated rationales boosts our local model's summarization performance.
\item We enhance model interpretability by analyzing LLM-derived rationales, deepening our insight into their summarization processes.
\end{itemize}

Overall, our study streamlines powerful summarization models in resource-limited contexts, offering insights into harnessing LLMs' inherent summarization abilities.

\section{Related Work}

\noindent\textbf{Text Summarization using LLMs}.
Transformer-based language models \citep{vaswani2017attention} have improved the quality of text summarization significantly.  These models excel at capturing complex relationships in long texts. Recent research has taken this transformer architecture further for summarization tasks~\citep{liu2019text, lewis2019bart, zhang2020pegasus, raffel2020exploring},  utilizing LLMs such as ChatGPT, GPT-4, and PaLM \cite{openai2023gpt4, chowdhery2022palm} which have billions of parameters and are trained on vast amounts of text. Their performance can be further enhanced when prompted to execute step-by-step reasoning \cite{wei2023chainofthought}.

However, the resource demands of LLMs have limited their widespread use. Concerns over privacy when using LLM-as-a-service APIs have also arisen, especially for sensitive data. This highlights the need for more compact local models that can still capture summarization abilities.
To harness the summarization ability of LLMs, \citet{wang-etal-2021-want-reduce} uses LLMs to augment labels for headline generation, while \citet{liu2023learning} used summaries created by LLMs as benchmarks for training their local models. LLMs were also used to evaluate summary quality during training. However, this approach did not fully transfer the reasoning skills of LLMs to the local models, indicating a partial capture of LLMs' summarization abilities. Also, the uncertainty of labels generated by deep learning models may affect reliability. \\

\noindent\textbf{Rationale Distillation for Interpretability in LLMs}
Knowledge distillation, as introduced  by \citet{hinton2015distilling}, refers to the concept for transferring knowledge from a large model (teacher) to a smaller one (student)  to make deep learning models usable in resource-limited environments. This idea has been applied and extended across various fields~\citep{sanh2019distilbert, tang2019distilling, jiao2019tinybert, chen2019distilling, lin2020knowledge, wang2023anypredict}. 
Notably, \citet{chen2019distilling} focused on abstractive summarization, while \citet{lin2020knowledge} emphasized extractive summarization. The complexity of deep neural networks has driven research toward making AI models interpretable~\citep{ribeiro2016should, doshi2017towards}. Rationale generation is an emerging technique in interpretability, highlighting a model's key reasoning steps~\citep{zaidan2008modeling, yu2020reclor}. In knowledge distillation, rationale generation enhances interpretability, offering insights into the  decision-making of LLMs. This informs the development of better knowledge distillation methods.  \citep{wang2022pinto} developed a smaller model using LLM-generated rationales and questions. Others \citep{shridhar2023distilling, ho2023large, magister2023teaching, hsieh2023distilling} used LLM-produced rationales to train models, improving performance and transparency in predictions, primarily for tasks like QA, NLU, arithmetic reasoning, and extractive summarization~ \cite{yang2023exploring}. This has left a gap concerning abstractive text summarization. To bridge this gap, we introduce an \textit{aspect-triple} rationale generation approach, aimed at distilling the summarization prowess of LLMs. This method consists of a procedure of extracting essential aspects, pinpointing primary relationships, and constructing a definitive summary.

\begin{figure*}[!h]
    \centering
    \includegraphics[width=\linewidth]{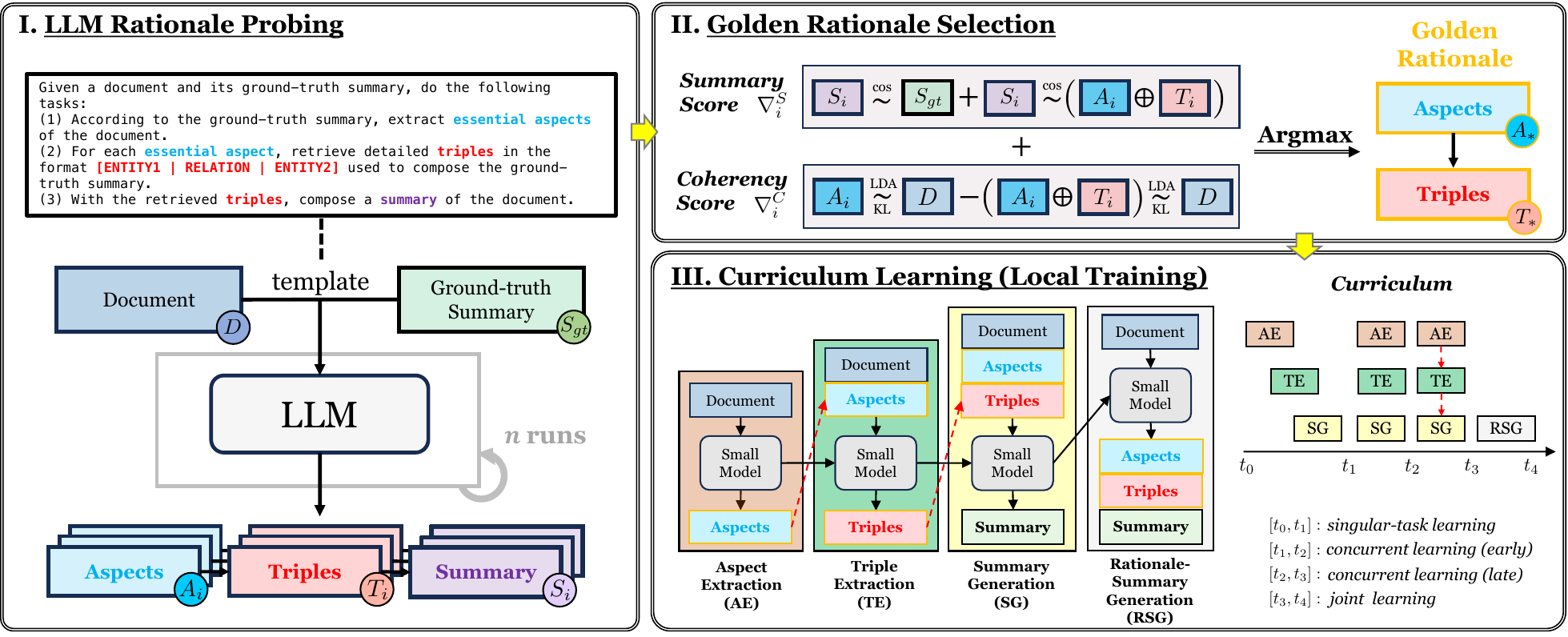}
    \caption{\textbf{Distilling text summarization ability from LLM to local model using \method}. \textbf{Step 1. LLM Rationale Probing: } Employing a template-based prompt incorporating the given document and ground-truth summary, we engage an LLM to generate a set of $n$ step-by-step rationales across $n$ iterations. \textbf{Step 2. Golden Rationale Selection:} We leverage summary and coherency scores to meticulously choose high-quality training rationales, enhancing the training dataset. \textbf{Step 3. Curriculum Learning:} We implement a curriculum learning strategy to train our compact small model with rationalized summarization ability from easy to challenging tasks.}
    \label{fig:framework}
\end{figure*}
\section{Method}
\subsection{Overview of \texttt{TriSum}}
We introduce \method, an approach transferring document summarization ability from an LLM ($\geq$100B) to a small LM ($\leq$1B) via rationale probing, golden rationale selection, and curriculum learning. Here, we assume the LLM has reasoning ability and can be used for prompting.
Before discussing in detail, we define a few key concepts and notations below.

\begin{definition} [\textbf{Aspect}]
    An (essential) aspect $\alpha$ is defined as a few words representing a distinct topic in a document.
\end{definition}
- \textit{Example}: \textit{
In a document about climate change, an aspect might be "rising sea levels".}

\begin{definition} [\textbf{Triple}]
    A triple $\tau =$ $\lb s | r | o\rb$ is a structure formatting a piece of free-text into a subject $s$, a relation $r$, and an object $o$.
\end{definition}
- \textit{Example}: \textit{For a sentence ``Cats eat fish.'', ``Cats'' is the subject, ``eat'' is the relation, and ``fish'' is the object, forming a triple $\lb Cats | eat | fish\rb$.}

\begin{task} [\textbf{Aspect Extraction (AE)}]
\label{task:asp_ext}
Given a document $D$, the task of aspect extraction is defined as extracting its essential aspects $A$ (where each $\alpha \in A $ represents an aspect) that approximates the distribution $p(A | D)$.   
\end{task}

\begin{task} [\textbf{Triple Extraction (TE)}]
\label{task:tri_ext}
Given a document $D$ and its aspects $A$, the triple extraction task is defined as extracting triples $T$ (where each $ \tau \in T $ represents a triple) from $D$, aiming to learn the distribution $p(T | D, A)$. 
\end{task}

\begin{task}[\textbf{Summary Generation (SG)}]
\label{task:sum_gen}
Given a document $D$, its aspect $A$, and the triples $T$, the task of summary generation is defined as generating a summary $S$ that approximates the distribution $p(S | D, A, T)$.
\end{task}

\begin{task}[\textbf{Rationale-Summary Generation (RSG)}]
\label{rat_sum_gen}
Given a document $D$, the task of rationale-summary generation is defined as generating both rationale and summary that approximates the distribution $p(A, T, S | D)$.
\end{task}

\noindent As illustrated in Figure \ref{fig:framework}, \method operates through three key steps: (1) tapping into the LLM for \textit{aspect-triple} rationales in training data; (2) selecting golden (high-quality) rationales based on summary and coherency scores; and (3) training a local model using a curriculum learning approach. We detail each step of \method as follows.

\subsection{Step 1: LLM Rationale Probing}

Given a set of documents for training, our initial step involves leveraging the LLM to iteratively generate a set of \textit{aspect-triple} rationales alongside their corresponding summaries. The objective is the following: first, to enable the LLM to pinpoint essential aspects, and subsequently, to elaborate on each aspect using detailed triples.

In this process, the auto-regressive LLM generates both the rationale \(R\) and the summary \(S\). We denote the length of a sequence by \(|\cdot|\). The rationale \(R = (A, T)\) is a sequence of tokens \(\{r_1, r_2, ..., r_{|R|}\}\), which is composed of aspect tokens \(\{a_1, a_2, ..., a_{|A|}\}\) followed by triple tokens \(\{t_1, t_2, ..., t_{|T|}\}\), where \(|R| = |A| + |T|\). Here, \(A\) represents essential aspects, and \(T\) provides detailed triples. Each \(a_i\) is an individual token in \(A\), and each \(t_j\) is an individual token in \(T\). The summary $S$ is defined as $\{s_1, s_2, ..., s_{|S|}\}$. Each token $r_i$ is generated based on the document $D$, the ground-truth summary $S_{gt}$, and the tokens previously generated, $R^{<i} = \{r_1, r_2, ..., r_{i-1}\}$. The prediction of $s_i$ is contingent upon the generated rationale $R$ and $S^{<i} = \{s_1, s_2, ..., s_{i-1}\}$:
\begin{equation}
\begin{aligned}
& p(R|D, S_{gt}) = \prod_{i=1}^{u} p(r_i|D, S_{gt}, R^{<i}), \\
& p(S|D, S_{gt}, R) = \prod_{i=1}^{v} p(s_i|D, S_{gt}, R, S^{<i}).
\end{aligned}
\end{equation}
where $S_{gt}$ denotes the ground-truth summary corresponding to the document $D$. To equip our local model with more interpretable and high-quality rationales, we prompt the LLM for $n$ iterations, which results in $n$ pairs of rationale-summary, denoted as $\{R_i, S_{i}\}^{n}_{i=1}$ for each document. Each pair, where $R_i = (A_i, T_i)$, serves as a candidate for the golden rationale selection described as follows.

\subsection{Step 2: Golden Rationale Selection}
Given the generated candidate rationales, we then incorporate two types of scores - \textit{Summary Score} and \textit{Latent Dirichlet Allocation (LDA)-based Coherence Score} to select the golden rationales.

\paragraph{Summary Score.} For each rationale $R_i$ in the candidates $\{R_i, S_{i}\}^{n}_{i=1}$, suppose $\hat{R_{i}}$, $\hat{S}_{i}$, and $\hat{S}_{gt}$ are the word embeddings of the rationale, LLM-generated summary, and the ground-truth summary respectively, the summary score is a weighted average of two semantic similarity:
\begin{align}
\label{eq:summary_score}
\nabla^{S}_i =  \mathrm{sim} \langle \hat{S}_{i}, \hat{S}_{gt} \rangle + \phi_{\alpha} \cdot \mathrm{sim} \langle \hat{S}_{i}, \hat{R_{i}} \rangle,
\end{align}
where $\phi_{\alpha}$ is a hyper-parameter balancing the importance of two components, and $\mathrm{sim}\langle\cdot\rangle$ is the semantic similarity computation. For example, $\mathrm{sim} \langle x,  y \rangle$ can be computed using cosine similarity as $\mathrm{sim} \langle x,  y \rangle = \frac { x \cdot y}{|| x|| \cdot || y||}$. The first term in Eq.~\eqref{eq:summary_score} emphasizes the similarity between the generated summary and the ground-truth summary, while the second term focus on the relevance between the generated summary and the prepended rationale, in avoid scoring high for lazy generation by the LLM (i.e., simply repeat the given ground-truth summary regardless of the generated rationale).

\paragraph{Coherence Score.} We also want to evaluate how the aspects and rationale align with the latent topics of the document. Here, we employ a Latent Dirichlet Allocation (LDA) model \cite{blei2003latent}, an algorithm that represents each document as a blend of a certain number of topics. To be specific, we represent each document as a distribution over the entire lexicon. Given a document $D$, a rationale $R_i$, and aspects $A_i \in R_i$, we initially train an LDA model on the corpus (all documents in the dataset) to identify latent topics with our specified number of topics $k$. It is important to clarify that the topics identified by LDA are based on the entire corpus, in contrast to the aspects which are specific to individual documents. From this model, we derive the topic distributions $p^D_{\mathrm{LDA}}$, $p^A_{i,\mathrm{LDA}}$, and $p^R_{i,\mathrm{LDA}}$ for the document, the $i$-th aspects, and the $i$-th rationale, respectively.
The coherence score $\nabla^{C}_i$ is calculated as the KL-divergence between these distributions:
\begin{align}
\label{eq:lda_score}
    &\nabla^{C}_i = KL(p^D_{\mathrm{LDA}} || p^A_{i,\mathrm{LDA}}) &\nonumber\\
    &\quad - (1 + \phi_{\beta}) \cdot KL(p^D_{\mathrm{LDA}} || p^R_{i,\mathrm{LDA}})
\end{align}
where $\phi_{\beta}$ is a parameter that manages the weight of the $KL(p^D_{\mathrm{LDA}} || p^R_{i,\mathrm{LDA}})$ term itself, and $KL(\cdot||\cdot)$ symbolizes the KL-divergence computation:

\noindent The score $\nabla^{C}_i$ in Eq. \eqref{eq:lda_score} fosters two primary objectives: (1) $-\phi_{\beta} \cdot KL(p^D_{\mathrm{LDA}} $ $|| p^R_{i,\mathrm{LDA}})$, an term that enhances the topical coherence between the document and rationale. (2) $KL(p^D_{\mathrm{LDA}} || p^A_{i,\mathrm{LDA}}) - KL(p^D_{\mathrm{LDA}} || p^R_{i,\mathrm{LDA}})$, a term which encourages the triples ($T_i \in R_i$) to refine this coherence beyond what is achieved by aspects alone.

The final selection of optimal rationales, denoted as $R_{*} = (A_{*}, T_{*})$, is based on those that yield the highest combined score of Eq. \eqref{eq:summary_score} and Eq. \eqref{eq:lda_score}, and given by Eq.~\eqref{eq:finals},
\begin{align}
R_{*} = \mathrm{argmax}_{i} (\nabla^{S}_i + \lambda_{cs} \cdot \nabla^{C}_i),
\label{eq:finals}
\end{align}
where $\lambda_{cs}$ is a balancing hyperparameter that manages the relative contributions of the two scores. We then use the gold rationales as the supervision to train our local lightweight language model in the following step.

\subsection{Step 3: Curriculum Learning}
To train the student Seq2Seq language model with the selected golden rationales for rationalized text summarization, we introduce an approach reminiscent of curriculum learning \cite{bengio2009curriculum, hacohen2019power, nagatsuka-etal-2021-pre, xu-etal-2020-curriculum}, which facilitates learning in stages of increasing complexity. This strategy consists of the following phases: (1) Singular-task learning, (2) Concurrent learning, and (3) Joint learning. For the first two phases, we focus on the tasks of \textit{aspect extraction}, \textit{triple extraction}, and \textit{summary generation}, distinguished by prefix tokens \texttt{\lb AspExt\rb}, \texttt{\lb TriExt\rb}, and \texttt{\lb SumGen\rb }, respectively. We use prefix tokens \texttt{\lb article\rb}, \texttt{\lb aspects\rb}, \texttt{\lb triples\rb}, \texttt{\lb summary\rb} to specify $D$, $A$, $T$, and $S$, respectively.

\paragraph{Singular-task learning}
Initially, we train the model on each task separately, aiding the model in developing a baseline understanding and ability to handle each task individually. For instance, in \textit{aspect extraction}, we aim to train a model that minimizes the loss $\mathcal{L}_{A}$ given the document $D$:
\begin{align}
\mathcal{L}_{A} = - \sum_{D \in \mathcal{D}} \log p(A_* | D; \theta_{s}), \nonumber
\end{align}
where $\mathcal{D}$ is the training set of documents,  $p(A | D) = \prod_{j=1}^{m} p(a_{j} | D, A^{<j})$, with $m$ the length of the aspects in the rationale, $a_{j}$ the $j$-th token of the aspects, and $A^{<j}$ the previous generated aspect tokens. The model follows a similar procedure for \textit{triple extraction} and \textit{summary generation}, focusing on minimizing losses $\mathcal{L}_{T}$ and $\mathcal{L}_{S}$, respectively:
\begin{align}
\mathcal{L}_{T} = - \sum_{D \in \mathcal{D}} \log p(T_* | D, A_*; \theta_{s}), \nonumber\\
\mathcal{L}_{S} = - \sum_{D \in \mathcal{D}} \log p(S_{gt} | D, A_*, T_*; \theta_{s}). \nonumber
\end{align}

\paragraph{Concurrent Learning}
Once the model has become proficient in performing individual tasks, we advance to the concurrent learning phase where the model simultaneously learns the tasks. This phase allows for task interplay and reciprocal reinforcement of learning. To facilitate a smooth transition, we further split this phase into early and late stages.

\noindent\textit{Early Stage: LLM-guided Training}.
In the early phase, we use the aspects $A_{*}$ and triples $T_{*}$ from the best rationale $R_{*}$, along with the document $D$, as the supervisory signal for each task. The model is trained to minimize the loss:
\begin{align}
\label{eq:loss_concurrent_early}
    &\mathcal{L}_{\mathrm{concurrent-early}} = - \sum_{D \in \mathcal{D}} \biggl[ \log p(A_* | D; \theta_{c})  &\nonumber\\
    & \quad + \log p(T_* | D, A_*; \theta_{c}) + \log p(S_{gt} | D, R_*; \theta_{c})   \biggr]. \nonumber
\end{align}
Using the LLM's output as a form of teacher forcing \cite{bengio2015scheduled} allows the model to focus on learning the structured (aspect-triple-summary) summarization in the early stage, without its own flawed prediction distracting it.

\noindent\textit{Late Stage: Self-guided Training}. As we transition to the later stages, our focus pivots to training the model using its own predictions as inputs for subsequent tasks. This strategy is characterized by a cascading training approach: the model begins with aspect extraction, progresses to triple extraction, and ultimately leads to summary generation. The benefit of this approach stems from its sequential information flow, where the outcome of one task informs the next. However, a challenge emerges due to the computational overhead of decoding intermediate results, such as aspects and triples. To mitigate this, while maintaining the sequential integrity, we employ greedy decoding. This method accelerates the process by selecting the most likely token at each step, eliminating the need for full-blown generation at every juncture.
Based on this, the loss becomes:
\vspace{-0.15em}
\begin{align}
    &\mathcal{L}_{\mathrm{concurrent-late}} = - \sum_{D \in \mathcal{D}} \biggl[ \log p(A_{*} | D; \theta_{c})  &\nonumber\\
    & + \log p(T_{*} | D, \Tilde{A}; \theta_{c}) + \log p(S_{gt} | D, \Tilde{A}, \Tilde{T}; \theta_{c}) \biggr], \nonumber
\end{align}
where $\Tilde{A}$ and $\Tilde{T}$ represent the intermediate aspects and triples obtained generated through greedy decoding by the model itself. The primary aim of this phase is twofold: (1) to diminish the model's dependency on LLM-provided rationales and, (2) to augment the model's capability for autonomous learning, with the overarching aspiration of enabling it to generate its own rationales and summaries.

\paragraph{Joint Learning}
In the final phase, we enhance the model's ability to concurrently generate both the rationale and the summary from a given document with the \textit{rationale-summary generation} task. Different from the late stage of concurrent learning, this stage streamlines the process by collapsing three pairs of encode-decode processes into a single pair. We use the optimal rationale from the LLM and the ground-truth summary as the labels. We introduce the prefix token \texttt{\lb RatGen\rb} for this task. The model aims to minimize the following loss function:
\vspace{-0.15em}
\begin{align}
    &\mathcal{L}_{\mathrm{joint}} = - \sum_{D \in \mathcal{D}} \biggl[ \lambda_{R} \log p(R_{*} | D; \theta_{r})  &\nonumber\\
    &\quad\quad\quad\quad\quad\quad\quad\quad  + \lambda_{S} \log p(S_{gt} | D, \Tilde{R}; \theta_{r}) \biggr], \nonumber
\end{align}
where $S_{gt}$ is the human-annotated ground-truth summary in the dataset, $\Tilde{R}$ is the generated rationale via greedy decoding, and $\lambda_{R}$ and $\lambda_{S}$ are hyperparameters that balance the importance of rationale and summary generations. 

Through our strategically designed curriculum learning process, the model progressively gains the capability to generate accurate and succinct rationales and summaries.

\begin{table}[!tp]
\small
\centering
\resizebox{\linewidth}{!}{
\begin{tabular}{lccccc}
\toprule
& \multicolumn{3}{c}{\textbf{\# Samples}} & \multicolumn{2}{c}{\textbf{\# Words}}   \\      
\textbf{Dataset} &Train &Valid &Test &Doc. &Sum. \\
\midrule
CNN/DailyMail &287,113 &13,368 &11,490 &766.6 &54.8 \\
XSum &204,045 &11,332 &11,334 &414.5 &23.0 \\
ClinicalTrial &163,088 &20,386 &20,386 &181.4 &45.2 \\
\bottomrule
\end{tabular}
}
\caption{\textbf{Statistics of datasets.}}
\vspace{-1em}

\label{tb:dataset_stat}
\end{table}
\section{Experiments}
\paragraph{Data Source}
Our evaluation of \texttt{TriSum} is carried out using three datasets: CNN/Daily Mail (CNNDM) v3.0.0 \citep{nallapati2016abstractive}, XSum \citep{narayan1808don}, and a bespoke dataset we have developed from Clinical Trial\footnote{\url{https://clinicaltrials.gov/}}. The comprehensive statistics of these datasets can be found in Table \ref{tb:dataset_stat}. To construct the ClinicalTrial dataset, we treat the "detailed description" from Clinical Trial as the document and the "brief summary" as its corresponding ground-truth summary. From an original total of 305,591 samples, we have selected 203,860 (with a splitting ratio of 8:1:1), filtering out entries where documents exceed 1,024 tokens or where summaries surpass 256 tokens. 

\paragraph{Model and Parameters} For the rationale generation and the summarization process, we employ GPT-3.5 (specifically, the gpt-3.5-turbo\footnote{We use the checkpoint gpt-3.5-turbo-0613, available at \url{https://platform.openai.com/docs/models/gpt-3-5}}) as the LLM. In the LLM rationale probing phase, we prompt the LLM differently for each dataset: $n=\{15, 8, 8\}$ times for CNNDM, XSum, and ClinicalTrial respectively. This generates a diverse set of potential rationale candidates. The parameters for the golden rationale selection are set as follows: $\phi_{\alpha} = 0.6$, $\phi_{\beta} = 1.3$, and $\lambda_{cs}=1.5$. We use cosine similarity to calculate the summary score with the embeddings retrieved from text-davinci-003 (a GPT-3.5 model that provides embedding). LDA latent topics are specified at 200, 500, and 300 for CNNDM, XSum, and ClinicalTrial respectively. For the joint learning phase, the parameters are fixed at $\lambda_{R} = 0.8$ and $\lambda_{S} = 1.2$.

\begin{table*}[!tp]
\small
\centering
\resizebox{\textwidth}{!}{
\begin{tabular}{lcccc|cccc|cccc}
\toprule
&\multicolumn{4}{c}{\textbf{CNN/DailyMail}} &\multicolumn{4}{c}{\textbf{XSum}} &\multicolumn{4}{c}{\textbf{ClinicalTrial}}   \\  
\midrule
\textbf{Model} & \textbf{R-1}   & \textbf{R-2}   & \textbf{R-L} &$\Delta$   & \textbf{R-1}   & \textbf{R-2}   & \textbf{R-L} &$\Delta$  & \textbf{R-1}   & \textbf{R-2}   & \textbf{R-L}   &$\Delta$      
\\    
\midrule
\textbf{Baselines} 
\\
BERTSumAbs \cite{liu2019text}
&41.2      
&18.7       
&37.2               
&$+$13.6\%      

&38.8       
&16.5       
&31.0      
&$+$28.3\%  

&39.2
&19.3
&29.6
&$+$19.3\%

\\

$\textrm{T5}_\textrm{Large}$ \textrm{\cite{raffel2020exploring}} 
&42.4       
&20.8       
&39.9               
&$+$7.0\%       

&40.1       
&17.2       
&32.3       
&$+$23.5\% 

&41.3
&22.1
&32.5
&$+$9.6\%

\\

\rowcolor{gray!20} ${\textrm{BART}_{\textrm{Large}}}$ \cite{lewis2019bart}
&44.0       
&21.1       
&40.6               
&$+$4.4\%       

&45.4       
&22.3       
&37.3       
&$+$5.4\%    

&43.5
&23.3
&33.7
&$+$4.6\%

\\

PEGASUS \cite{zhang2020pegasus} 
&44.2       
&21.6       
&41.3 
&$+$3.0\%     

&\textbf{46.7}  
&\textbf{24.4}  
&38.9 
&$+$0.6\% 

&41.8
&22.9
&31.7
&$+$9.0\%
\\

GSum \cite{dou-etal-2021-gsum} 
&45.5
&22.3
&\textbf{42.1} 
&$+$0.4\%  

&45.1
&21.5 
&36.6 
&$+$7.3\% 

&43.5
&23.1
&32.8
&$+$5.7\%  

\\

$\textrm{BigBird}_\textrm{Large}$ \textrm{\cite{zaheer2021big}} 
&43.8 
&21.1 
&40.7 
&$+$4.5\% 

&47.1
&24.1 
&38.8 
&$+$0.6\% 

&\textbf{44.2}
&23.8
&\textbf{34.5}
&$+$2.5\%

\\

SimCLS \cite{liu-liu-2021-simcls} 
&45.6
&21.9
&41.0
&$+$1.7\% 

&46.6
&\textbf{24.2}
&\textbf{39.1}
&$+$0.7\% 

&43.8
&23.3
&34.1
&$+$3.9\%

\\

SeqCo \cite{xu2022sequence} 
&45.0 
&21.8 
&41.8 
&$+$1.6\% 

&45.6 
&22.4 
&37.0 
&$+$5.4\% 

&42.8
&22.5
&33.2
&$+$6.7\%
\\

$\textrm{GLM}_\textrm{RoBERTa}$ \textrm{\cite{du-etal-2022-glm}}
&43.8 
&21.0 
&40.5 
&$+$4.7\%  

&45.5 
&23.5 
&37.3 
&$+$4.1\% 

&43.3
&23.0
&33.9
&$+$4.9\% 

\\

$\textrm{GPT-3.5}_\textrm{zero-shot}$
&37.4 
&13.8 
&29.1 
&$+$37.4\% 

&26.6 
&6.7 
&18.8
&$+$112.5\% 

&34.8
&12.8
&23.5
&$+$47.8\%

\\

\midrule
\textbf{Our Method} \\
GPT-3.5 w/ \texttt{TriSum} rationale 
&\textbf{46.7} 
&\textbf{23.5} 
&40.7 
&$-$0.5\% 

&34.4 
&12.6 
&28.4 
&$+$46.8\% 

&\textbf{44.6}
&\textbf{24.5}
&30.4
&$+$5.6\%

\\

\texttt{TriSum-S}
&\textbf{45.9} 
&\textbf{22.8} 
&\textbf{42.3} 
&$-$0.6\% 

&\textbf{47.4} 
&\textbf{24.8} 
&\textbf{39.4} 
&$-$1.0\%

&\textbf{45.3}
&\textbf{24.8}
&\textbf{35.0}
&$+$0.0\%

\\

\texttt{TriSum-C}  
&45.5 
&22.3
&41.2 
&$+$1.2\% 

&46.5 
&24.0 
&38.7 
&$+$1.1\% 

&\textbf{44.2}
&23.7
&34.4
&$+$2.7\%
        
\\

\texttt{TriSum-J}
&\textbf{45.7} 
&\textbf{22.7} 
&\textbf{41.9} 
& ---

&\textbf{47.3} 
&\textbf{24.4} 
&\textbf{39.0} 
& ---

&\textbf{45.3}
&\textbf{24.6}
&\textbf{35.2}
& ---

\\
\bottomrule
\end{tabular}
}

\caption{\textbf{Performance comparison of ROUGE Scores across CNN/DailyMail, XSum, and ClinicalTrial datasets}. The labels \texttt{TriSum-S}, \texttt{TriSum-C}, and \texttt{TriSum-J} signify model checkpoints at the end of singular-task, concurrent, and joint learning stages, respectively. For \texttt{TriSum-S}, distinct optimal checkpoints, each tailored for a specific task, are used in a pipeline of three Seq2Seq models. The symbol $\Delta$ signifies the percentage improvement in the aggregate ROUGE scores achieved by \texttt{TriSum-J}. The top-3 results are \textbf{highlighted}. Our backbone model ${\textrm{BART}_{\textrm{Large}}}$ is shaded for reference. }
\vspace{-1em}
\label{tb:perf_all}
\end{table*}
\begin{table}[!h]
\small
\centering
\resizebox{0.48\textwidth}{!}{
\begin{tabular}{lcc|cc|cc}
\toprule
  &\multicolumn{2}{c}{\textbf{CNN/DailyMail}} &\multicolumn{2}{c}{\textbf{XSum}} &\multicolumn{2}{c}{\textbf{ClinicalTrial}}    \\  
\midrule
\textbf{Model} & \textbf{BS}   & \textbf{BAS}    & \textbf{BS}   & \textbf{BAS} & \textbf{BS}   &\textbf{BAS}       \\    
\midrule
\textbf{Baselines} \\
$\textrm{BERTSumAbs}$ &85.76 &-3.81 &87.23 &-3.66 &85.41 &-3.79 \\
${\textrm{T5}_{\textrm{Large}}}$  &87.22 &-3.71  & 90.73 &-2.70 &87.76 &-2.89\\
\rowcolor{gray!20}${\textrm{BART}_{\textrm{Large}}}$  &87.98 &-3.45  &91.62 &-2.50 & 88.30 & -2.79 \\
$\textrm{PEGASUS}$  &87.37 &-3.64  &91.90 &-2.44 &87.62 &-2.80 \\ 
$\textrm{GSum}$ &87.83 &-3.54 &91.23 &-2.57 &88.41 &-2.75\\ 
$\textrm{BigBird}_\textrm{Large}$  &88.03 &-3.38 &\textbf{91.97} &\textbf{-2.40}  &\textbf{89.45}  &-2.67  \\ 
$\textrm{SimCLS}$  &88.28 & -3.39  &90.78 &-2.93 &87.85 &-3.15 \\ 
$\textrm{SeqCo}$ &87.47 &-3.56 &91.35 &-2.56 &88.06 &-2.93 \\
$\textrm{GLM}_\textrm{RoBERTa}$  &87.33 &-3.69 &91.87 &-2.51 &88.55 &-2.84\\ 
$\textrm{GPT-3.5}_{\textrm{zero-shot}}$ &87.70 &-3.36 &87.67 &-2.80 &87.08 &-3.01\\
\midrule
\textbf{Our Method} \\
$\textrm{GPT-3.5}_\textrm{TriSum}^{*}$  &\textbf{89.20} &\textbf{-3.14}  &89.25 &-2.58 &89.20 &\textbf{-2.55}\\
$\texttt{TriSum-S}$  &\textbf{88.48} &\textbf{-3.22} &\textbf{91.95} &\textbf{-2.38} &\textbf{90.05} &\textbf{-2.47}\\
$\texttt{TriSum-C}$  &87.21&-3.76 &90.88 &-2.84 &89.40 &-2.59 \\
$\texttt{TriSum-J}$  &\textbf{88.50} &\textbf{-3.25} &\textbf{92.17} &\textbf{-2.33} &\textbf{89.97} &\textbf{-2.53}\\
\bottomrule
\end{tabular}
}

\caption{\textbf{Pre-trained language model-evaluated semantic similarity scores}. ``*'' indicate the inference with \method-generated rationale. ``BS'' and ``BAS'' are BERTScore and BARTScore, respectively. Top-3 results are \textbf{highlighted}.}

\vspace{-1em}
\label{tb:perf_sem_sim}
\end{table}
\paragraph{Training} For both CNNDM and XSum datasets, we utilize the BART-Large \cite{lewis2019bart} checkpoints
that have been fine-tuned specifically for these datasets, as the backbone models. In the case of ClinicalTrial, we fine-tune the BART-Large CNNDM checkpoint using only the summary to create a backbone model. All models, including the baselines, undergo fine-tuning for three epochs, with an early stopping mechanism in place to optimize performance. We train models with an NVIDIA RTX A6000 GPU.
 
\paragraph{Baselines} We compare \method to baseline abstractive summarization models including BERTSumAbs~\cite{liu-lapata-2019-text}, T5~\cite{raffel2020exploring}, BART~\cite{lewis2019bart}, PEGASUS~\cite{zhang2020pegasus},
GSum~\cite{dou-etal-2021-gsum}, BigBird~\cite{zaheer2021big}, SimCLS~\cite{liu-liu-2021-simcls}, SeqCo~\cite{xu2022sequence}, GLM~\cite{du-etal-2022-glm}, 
and GPT-3.5.

\paragraph{Evaluation}
We use the following metrics: (1) ROUGE-F1: measures the overlap of n-grams 
between the generated summary and the reference summary. We measure ROUGE-1 (R-1), ROUGE-2 (R-2), and ROUGE-L (R-L). (2) BERTScore and BARTScore: measure the semantic similarity between the generated summary and the reference summary using pre-trained language models $\textrm{RoBERTa}_{\textrm{Large}}$ and $\textrm{BART}_{\textrm{Large}}$, respectively.

\subsection{Performance Analysis}

Tables \ref{tb:perf_all} and \ref{tb:perf_sem_sim} provide an in-depth look at how our \method approach performs compared to various baseline models. The results include both ROUGE scores and semantic similarity metrics across different datasets, from general news sources to specialized domain-specific collections. Our analysis reveals several key insights:

\paragraph{Consistent Edge Over Baselines} The \method approach consistently outperforms many state-of-the-art models across different datasets, highlighting its strength and adaptability. Statistically, in terms of overall ROUGE scores, \texttt{TriSum-J} outperforms fine-tuned models (excluding GPT-3.5) by $4.5\%$ on CNNDM, $8.5\%$ on XSum, and $7.4\%$ on ClinicalTrial.
\begin{figure*}[t]
    \centering
    \includegraphics[width=\linewidth]{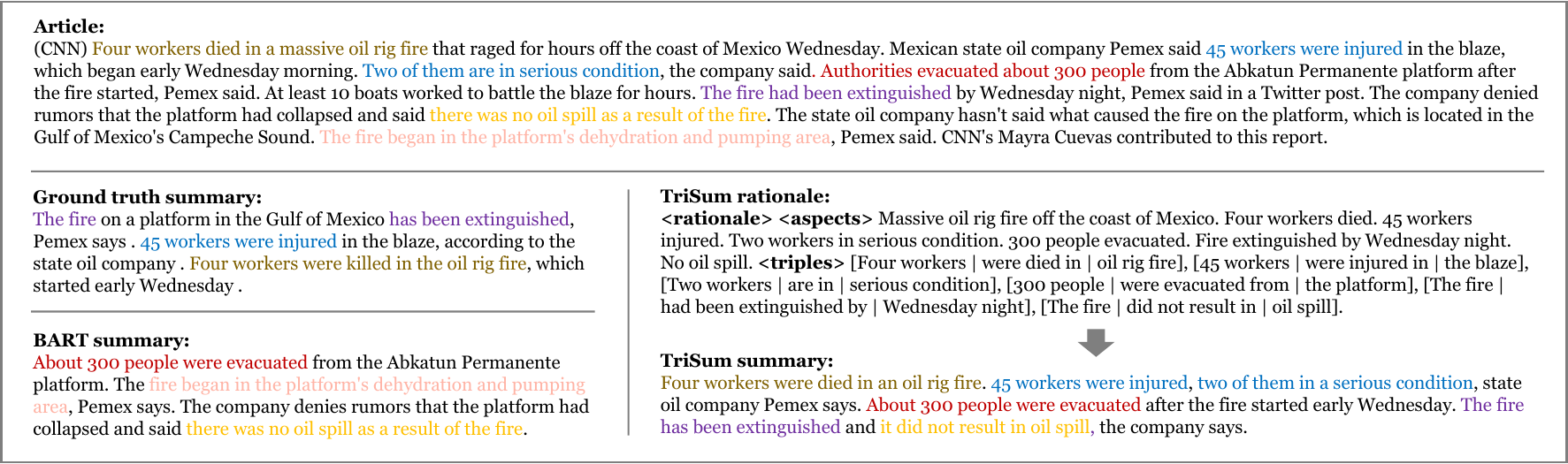}
    \caption{\textbf{An example of abstractive summarization on CNN/DailyMail dataset.} We compare the summary generated by our \texttt{TriSum} approach to the ground-truth summary and the one generated by BART. We use different colors to show the distinct topics in the article and summary.}
    \label{fig:case_study}
    \vspace{-1em}
\end{figure*}
\begin{figure}[!h]
    \centering
    \includegraphics[width=1.02\linewidth]{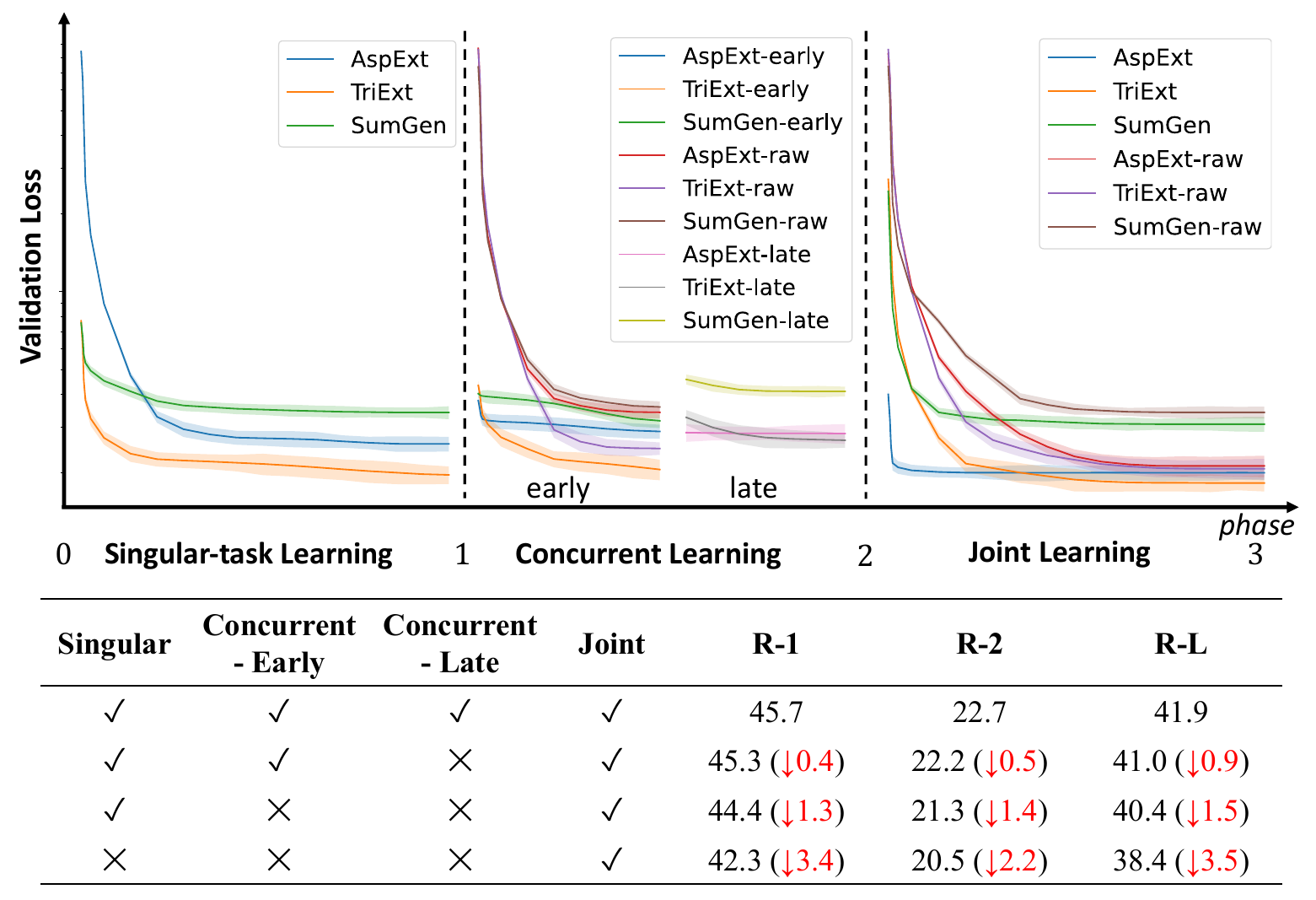}
    \caption{\textbf{Validation loss by training steps and ablation study for curriculum learning on CNN/DailyMail.} \textit{AspExt}, \textit{TriExt}, and \textit{SumGen} denote aspect extraction, triple extraction, and summary generation tasks, respectively. \textit{-early}/\textit{-late} denote the early/late stage of concurrent learning. \textit{-raw} denotes training the model from scratch.}
    \label{fig:curriculum}
    \vspace{-0.5em}
\end{figure}
\paragraph{Gains Over Backbone} We use BART as the backbone model, which is already known for its performance in summarization tasks. The noticeable overall improvement across all datasets ($+4.8\%$ ROUGE score and $+1.0\%$ BERTScore, and $+7.3\%$ BARTScore) when using the \method approach over BART is significant. This shows the effectiveness of including the LLM-generated rationales as the additional supervision and indicates the potential of our method to be scaled for the enhancement of other summarization models as well. Notably, \texttt{TriSum-S} consistently excels in performance. This heightened effectiveness is rooted in its modular design, which encompasses three checkpoints, each optimized for a unique task. Therefore, the improved results may be attributed to its thrice-enlarged parameter set, when compared to \texttt{TriSum-C} or \texttt{TriSum-J}.

\paragraph{Optimized Rationale for LLM} Interestingly, the rationales generated by \method can significantly improve the performance of GPT-3.5 within the dataset ($+40.9\%$ ROUGE Score, $+2.0\%$ BERTScore, and $+9.9\%$ BARTScore compared to $\textrm{GPT-3.5}_{\textrm{zero-shot}}$). For example, in our tests with the CNNDM dataset, the LLM, guided by the \method's rationale and without any fine-tuning, outperform all the other fine-tuned models in terms of ROUGE-1 score. This suggests that users can use fine-tuned \method to guide the LLM in creating quality summaries.

\paragraph{Effect of Curriculum Learning}
Figure \ref{fig:curriculum} shows the benefits of curriculum learning on the model's task performance. Two key comparisons are evident: the raw model versus one trained with singular-task learning in the early concurrent learning stage, and the raw model versus one trained through the previous two learning stages. The ablation study further reveals a step-wise performance improvement. Notably, when trained solely on joint learning from scratch, the model underperforms the original BART. This emphasizes the indispensable role of foundational tasks, without which BART struggles with the rationale-summary generation.

\begin{figure}[!tp]
\vspace{-0.5em}
    \centering
    \includegraphics[width=\linewidth]{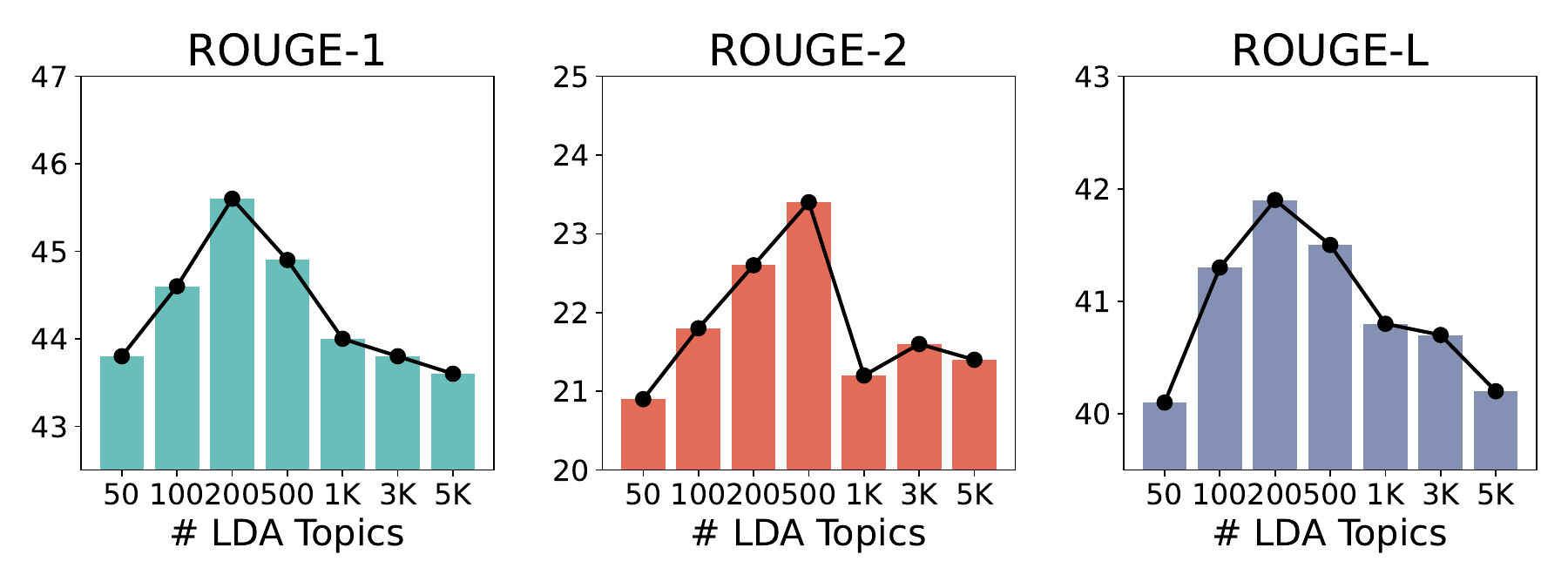}
    \caption{\textbf{Performance by different numbers of LDA latent topics specified in golden rationale selection.} We compare the ROUGE scores of the summaries generated by \texttt{TriSum-R} on CNN/DailyMail dataset.}
    \label{fig:perf_num_topics}
    \vspace{-0.5em}
\end{figure}

\paragraph{Effect of Golden Rationale Selection}
Figure \ref{fig:perf_num_topics} demonstrates the impact of our golden rationale selection. The performance of the trained model drops significantly when the number of latent topics is either too low (e.g., 50) or high (e.g., 5000). On the other hand, choosing an appropriate number of topics (e.g., 200) leads to improved outcomes. This underscores the importance of the quality of rationales; poor-quality rationales can negatively impact the model, emphasizing the value of our rationale selection strategy.

\paragraph{Case Study}
Figure \ref{fig:case_study} compares summaries created from a CNN article discussing an oil rig fire in Mexico. The ground truth summary adeptly encapsulates the main events, emphasizing the aftermath in terms of fatalities, injuries, and containment. BART's rendition, while detailed about the evacuation and fire's origin, misses out on pivotal information like the death toll and injury scale. On the other hand, \method's rationale begins by itemizing the essential aspects of the incident. These aspects present a high-level overview of the events and their aftermath. Following these aspects, the triples zoom into the specifics, elucidating the relations between the entities involved. This technique used by \method ensures a comprehensive summary and improves clarity. Readers can follow the summary's content back to its main aspects and detailed triples, gaining a deeper understanding of how the summarization process works. This transparency is a key feature of \method, allowing users to grasp the reasoning behind the summarized content. We provide more examples in the Appendix.



\section{Conclusion}
We introduced \method, an approach aimed at distilling summarization capabilities from a large language model to a small local model. Extensive experiments verified its superior performance over state-of-the-art models across diverse datasets on the abstractive summarization task. Our work highlights the potential of leveraging large model insights for efficient and nuanced text summarization.

\bibliography{references}

\clearpage
\appendix
\section{Ethics, Limitations, and Risks}

\subsection{Ethics}
\noindent \textbf{Data Privacy and Source:} All datasets used in this research, namely CNN/DailyMail, XSum, and ClinicalTrial, are publicly available\footnote{\url{https://github.com/abisee/cnn-dailymail}}\footnote{\url{https://github.com/EdinburghNLP/XSum}}\footnote{\url{https://clinicaltrials.gov/}}. This transparency minimizes ethical concerns related to data sourcing and usage.

\noindent \textbf{Interpretability:} The transparency and interpretability of AI models are ethical imperatives in many applications. \method not only improves summarization performance but also enhances the interpretability of the summarization process, making it more trustworthy.

\subsection{Limitations}
\noindent \textbf{Dependence on LLMs:} \method's effectiveness is contingent on the quality and capabilities of the LLMs it distills from. If the LLM has biases or inaccuracies, these could potentially be transferred to the local model.

\noindent \textbf{Scope of Rationales:} The \textit{aspect-triple} rationales, while enhancing interpretability, might not capture all nuances of the original text. Some information might be lost or oversimplified during the distillation process.

\subsection{Risks}
\noindent \textbf{Overfitting:} There's a potential risk that the local model might overfit to the rationales and summaries derived from the LLM, leading to reduced generalization on unseen data.

\noindent \textbf{Misinterpretation:} Enhanced interpretability can sometimes lead users to place undue trust in the model's outputs. Users should be cautious and consider the model's outputs as one of many tools in decision-making processes.

\noindent \textbf{Ethical Misuse:} Like all summarization tools, there's a risk that users might misuse \method to misrepresent complex information, leading to misinformation.

\section{Templates Used for Prompting LLM}
\label{ap:prompts}
In this section, we showcase the templates we used for prompting the large language model for different purposes. 
\begin{figure}[!h]
    \centering
    \includegraphics[width=\linewidth]{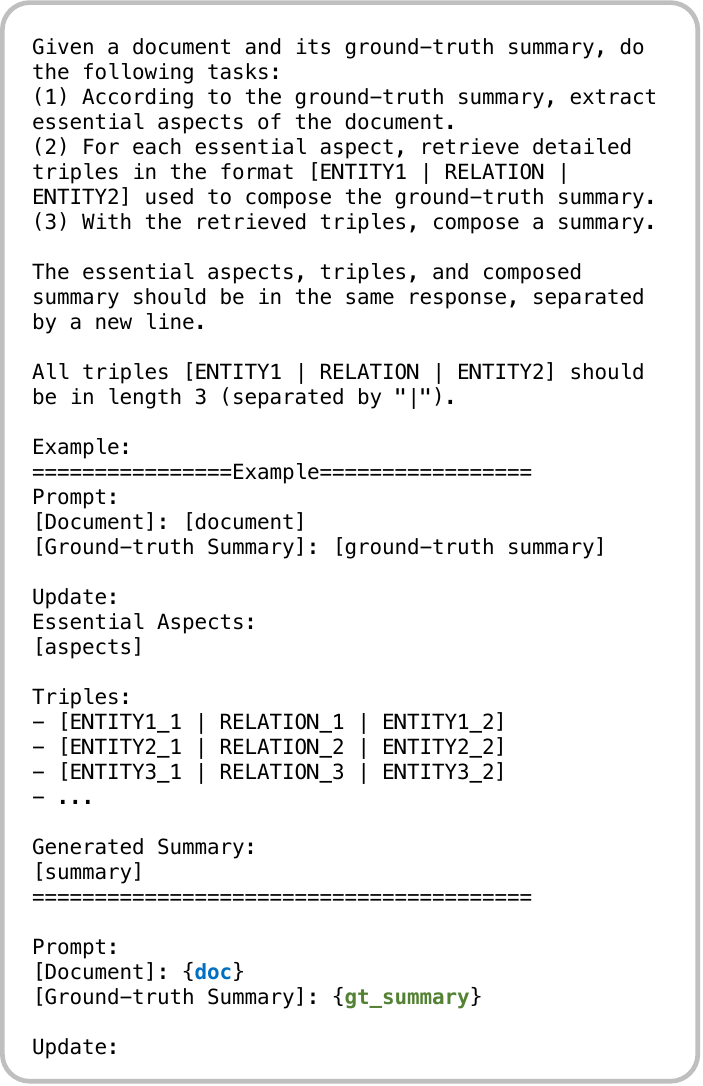}
    \caption{Template used for prompting rationale and summary from LLM}
    \label{fig:prompt_doc_gt}
\end{figure}
\begin{figure}[!h]
    \centering
    \includegraphics[width=\linewidth]{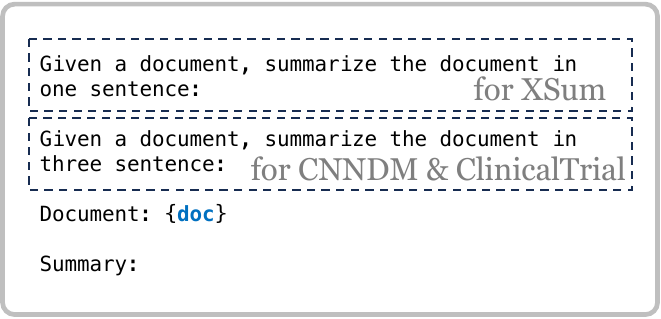}
    \caption{Template used for prompting summary from LLM in zero-shot setting.}
    \label{fig:prompt_zeroshot}
    \vspace{-0.5em}
\end{figure}
\begin{figure}[!h]
    \centering
    \includegraphics[width=\linewidth]{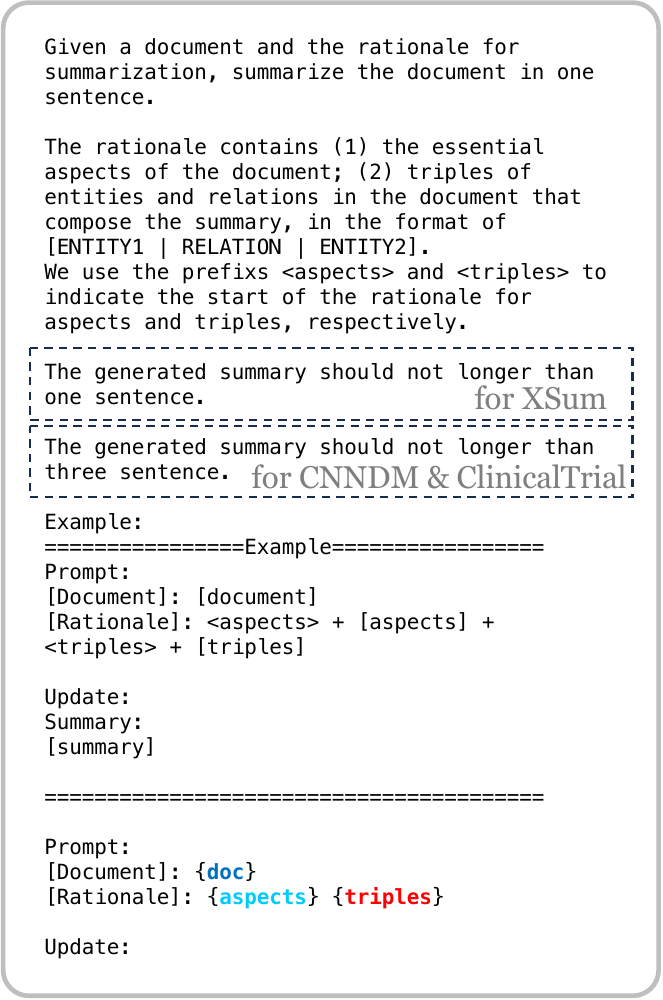}
    \caption{Template used for prompting summary from LLM given \method-generated rationale ($\textrm{GPT-3.5}_\textrm{TriSum}$).}
    \vspace{-0.5em}
    \label{fig:prompt_gpt_trisum}
\end{figure}

\noindent Figure \ref{fig:prompt_doc_gt} shows the template we use for \textbf{Step 1} (LLM Rationale Probing). It instructs the LLM to (1) generate essential aspects of the document with respect to the ground-truth summary; (2) extract triples from the document that elaborate on these key aspects; (3) generate a summary referring to both the retrieved triples and the ground-truth summary. The template then instructs the LLM to generate in a specific format, to reduce the randomness of the LLM's output. The document and the ground-truth summary are input to the placeholders to finalize the prompting request.

\noindent Figures \ref{fig:prompt_zeroshot} and \ref{fig:prompt_gpt_trisum} show the templates we use for testing the LLM's summarization ability in a zero-shot setting and with \method-generated rationales, respectively.

\section{Dataset Description}
\noindent \textbf{CNN/DailyMail} The CNN/DailyMail dataset is one of the most popular datasets for extractive and abstractive summarization tasks. Originating from online news stories, the dataset comprises articles from CNN and DailyMail websites. The overview of this dataset is described as follows:
\begin{itemize}[leftmargin=*]
    \item \textbf{Size}: It contains 287,113 training examples, 13,368 validation examples, and 11,490 test examples.
    \item \textbf{Content}: Each example in the dataset consists of a news article and several accompanying highlight points, which, when combined, form a coherent summary of the main article.
    \item \textbf{Nature of Summaries}: The highlights, crafted to engage a reader's attention, effectively form summaries. Typically, a summary consists of 2 to 3 sentences. They can be approached either extractively or abstractively by summarization models.
    \item \textbf{Usage}: Due to its substantial size and real-world data, CNN/DailyMail has been a benchmark for several state-of-the-art summarization models, enabling researchers to compare performances and strategies across diverse methods.
\end{itemize}

\noindent \textbf{XSum} XSum (Extreme Summarization) dataset provides a more challenging scenario for abstractive summarization. The overview of this dataset is described as follows:
\begin{itemize}[leftmargin=*]
    \item \textbf{Size}: It contains 204,045 training examples, 11,332 validation examples, and 11,334 test examples, which are the articles collected from the BBC (British Broadcasting Corporation).
    \item \textbf{Content}: Unlike CNN/DailyMail where summaries are constructed from highlights, each article in the XSum dataset is paired with a single-sentence summary, often written in a style that is not present in the article body.
    \item \textbf{Nature of Summaries}: The summaries in XSum are more abstractive in nature and are not simply extractive snippets from the articles. This demands models to truly understand the content and generate a unique summarizing sentence, making it a challenging dataset for abstractive summarization.
    \item \textbf{Usage}: XSum's distinctive nature has made it a preferred choice for researchers focusing on advanced abstractive methods in summarization. Its summaries, being creatively crafted and not directly extracted from the text, test the genuine abstracting capabilities of models.
\end{itemize}

\noindent\textbf{ClinicalTrial}
We collected the clinical trial protocol documents from clinicaltrials.gov where there are over 400K registered clinical trials across the world. The overview of this dataset is described as follows:
\begin{itemize}[leftmargin=*]
    \item \textbf{Size}: We downloaded the static copy of the whole clinical trial database which is with around 460K clinical trial documents. 203,860 were selected out of all based on the standard (a) they are interventional clinical trials, (b) missing or duplicate titles, (c) missing the brief summary section. To fit the context window of used language models, we further exclude documents that have more than 1024 tokens or the target summaries are with more than 256 tokens.
    \item \textbf{Content}: The clinical trial document describes the proposal for testing the effectiveness and the safety of a new treatment, e.g., a drug. The researchers need to list all the main elements required for FDA regulation, such as the title, proposed treatment, target condition, primary outcome measurements, eligibility criteria, etc. 
    \item \textbf{Nature of Summaries}: An effective summary of clinical trials need to deliver the main message about the motivation of the study as well as the route planning to reach the target. To make a good summary of clinical trials, the model needs a comprehensive view of the whole documents and maintain the key information.
    \item \textbf{Usage}: We will use the ``brief summary" section written by human experts provided in the raw clinical trial documents as the target for all models. 
\end{itemize}

\section{Interpretability of \method}
\begin{figure}[!h]
    \centering
    \includegraphics[width=\linewidth]{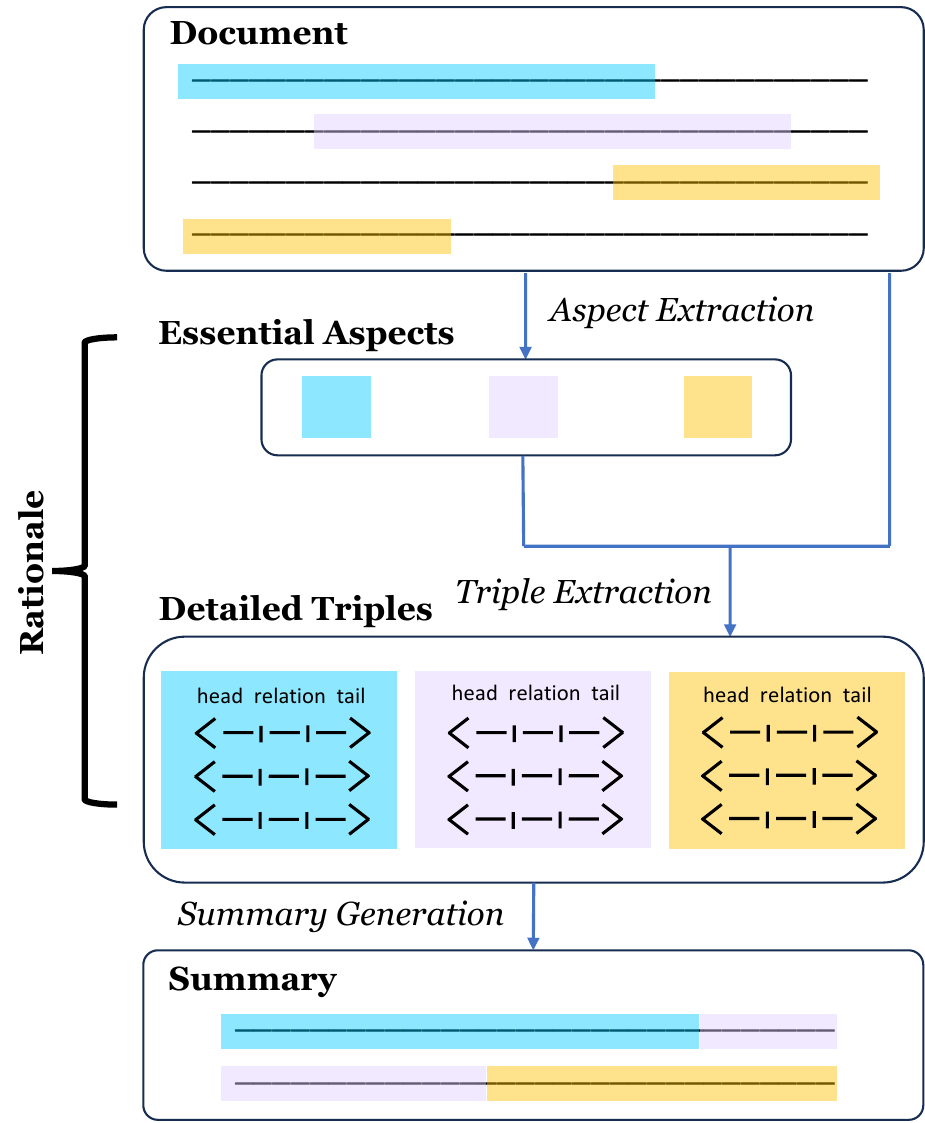}
    \caption{\textbf{Abstractive summarization with \method}. Different colors indicate different essential aspects covered by the document. We showcase how an \textit{aspect-triple} rationale is extracted and contribute to the final summary generation.}
    \label{fig:interpret}
\end{figure}

\noindent Interpretability is paramount in understanding and trusting AI systems, especially in tasks like abstractive summarization where the derivation of conclusions isn't always overtly apparent. The workflow of \method, illustrated in Figure \ref{fig:interpret}, is designed with this transparency in mind.

\begin{figure*}[!t]
    \centering
    \includegraphics[width=\linewidth]{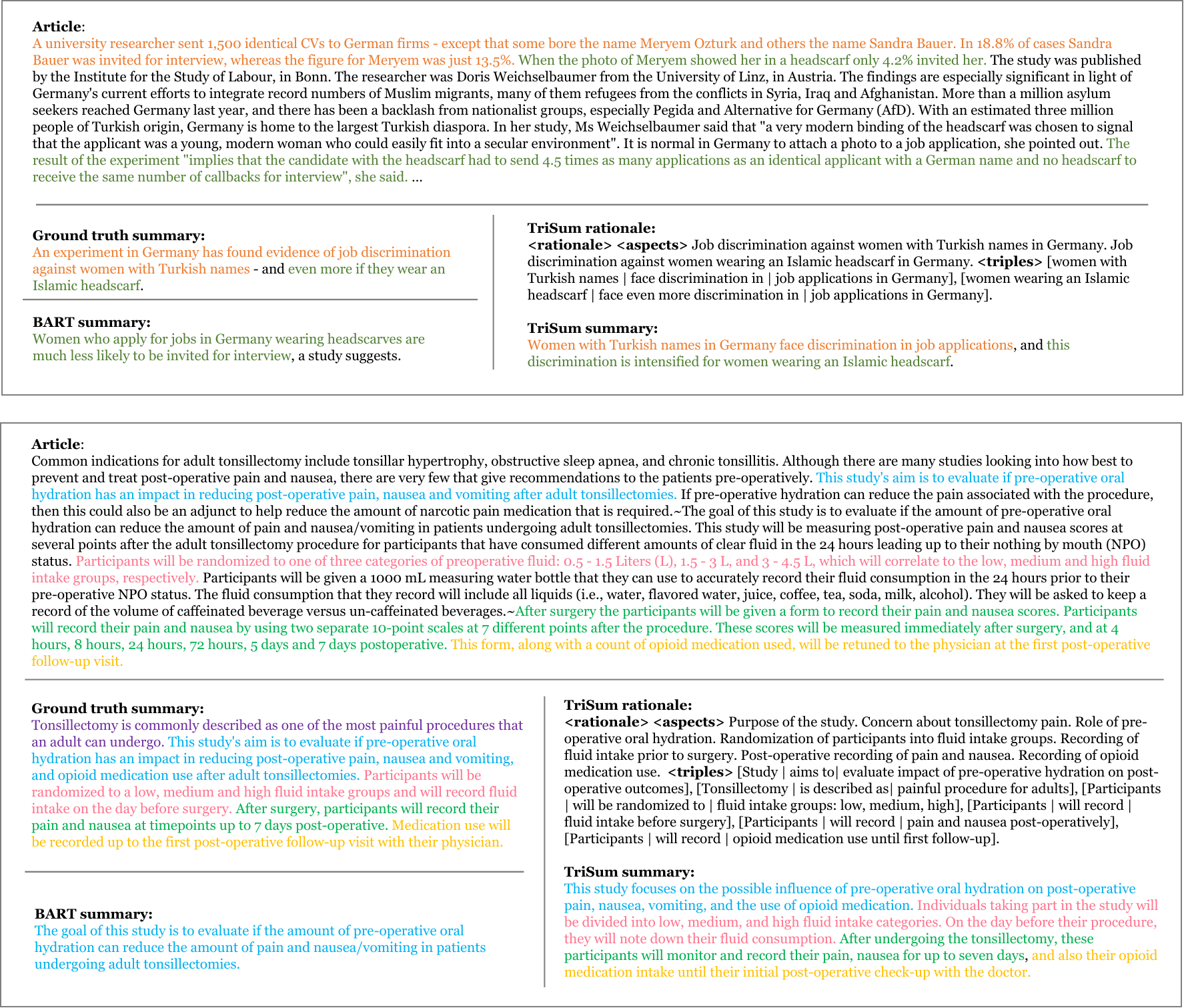}
    \caption{\textbf{Examples of abstractive summarization on XSum (above) and ClinicalTrial (below) datasets.} We compare the summary generated by our \texttt{TriSum} approach to the ground-truth summary and the one generated by BART. We use different colors to show the distinct topics in the article and summary.}
    \label{fig:case_study_appendix}
\end{figure*}
Starting with a given document, \method identifies its essential aspects. This step offers a clear insight into what the model perceives as the primary themes or topics within the document. Subsequently, using these aspects as anchors, \method revisits the document to meticulously extract triples, structured as \lb head | relation | tail\rb, for each aspect. These triples provide a structured, detailed representation, offering granular insights into the model's understanding of the relationships and entities in the text. Finally, \method fuses these extracted aspects and triples to produce a summary. By correlating the final summary with the previously identified aspects and triples, users can trace back the origins of particular summary fragments, gaining a clear understanding of how \method processes and abstracts information.

This step-by-step elucidation of the summarization process significantly enhances the model's transparency, making its decision-making rationale more discernible and hence fostering trust among its users.

\section{Hyperparameter Tuning}
\begin{table}[!h]
\small
\centering
\resizebox{\linewidth}{!}{
\begin{tabular}{cl}
\toprule
\multicolumn{1}{c}{\textbf{Hyperparameter}} & \textbf{Values} \\
\midrule
\multicolumn{1}{l}{\textbf{Golden Rationale Selection}}\\
$\phi_{\alpha}$ & \{0.2, 0.4, \textbf{0.6}, 0.8, 1.0, 1.2\} \\
$\phi_{\beta}$  & \{0.4, 0.6, 0.8, 1.0, \textbf{1.3}, 1.5, 2.0\}\\
$\lambda_{cs}$  & \{0.5, 1.0, \textbf{1.5}, 2.0 \} \\
LDA latent topics & \{50, 100, \textbf{200}, \textbf{300}, \textbf{500}, 1000, 3000, 5000\} \\ 

\multicolumn{1}{l}{\textbf{Rationale Learning}}\\
$(\lambda_{R}, \lambda_{S})$  & \{(1.0, 1.0), \textbf{(0.8, 1.2)}, (0.5, 1.5), (0.3, 1.7)\} \\

\bottomrule
\end{tabular}
}
\caption{\textbf{Hyperparameters of \method we tuned.} We highlight the optimal ones based on our experiments in \textbf{bold}.}

\label{tb:hyper}
\end{table}
Table \ref{tb:hyper} shows our comprehensive hyperparameter study to select the optimal values for \method.

\section{Case Studies}
In addition to Figure \ref{fig:case_study}, Figure \ref{fig:case_study_appendix} shows other two examples comparing our \method's performance with our backbone model BART on XSum and ClinicalTrial datasets. We can draw the following findings:

\subsection{Case Study on XSum}
In the given example, we juxtapose the performance of our approach, \method, with BART, our backbone model. Upon scrutinizing the sourced article detailing a research study on job discrimination against women with Turkish names and those wearing Islamic headscarves in Germany, we discern distinct nuances in the summaries rendered by both methods.

BART's summary encapsulates a broad understanding, highlighting that women wearing headscarves in Germany are at a disadvantage during job applications. While it successfully conveys a salient point, it omits the specific discrimination against women with Turkish names.

\method, on the other hand, demonstrates its prowess through a more holistic, nuanced, and detailed summary. It distinctly notes both aspects of the discrimination: one against women with Turkish names and the other against those donning an Islamic headscarf. \method's rationale section further accentuates its strength by explicitly presenting the core aspects and triples that delineate the focus points of the summary. This methodical extraction and representation ensure that no vital information is sidestepped.

Moreover, \method's summary doesn't merely report the findings but emphasizes the intensification of discrimination when both factors - a Turkish name and an Islamic headscarf - are combined. Such a layered insight is invaluable, especially in sensitive subjects such as discrimination, where capturing the entire scope of the issue is crucial.

In essence, while BART gives a generalized overview, \method offers a richer, more comprehensive narrative that mirrors the depth and breadth of the original article, underscoring the strength and precision of our approach.

\subsection{Case Study on ClinicalTrial}
In this case study centered around adult tonsillectomies, it is evident that the BART primarily grasped the core goal of the study but missed out on essential details, particularly the varied fluid intake groups and post-operative data recording. Meanwhile, the ground truth summary offers a comprehensive view, but it remains relatively generalized.

The strength of our approach, the \textit{aspect-triple} rationaled summarization (\method), is significantly highlighted when we delve into the details and the rationale-driven structure it adheres to. \method operates by identifying essential aspects of the text, followed by extracting and constructing triples that map the relationships in the content.
\begin{itemize}[leftmargin=*]
    \item \textbf{Aspect-Driven Understanding}: \method's rationale points out the key aspects such as the purpose of the study, concerns related to tonsillectomy pain, the role of pre-operative hydration, among others. By capturing these aspects, the model sets the stage for a summary that does not miss out on the diverse elements of the original text.
    \item \textbf{Triple-Based Detail Extraction}: The aspect-driven approach is further enriched by the triples \method generates. These triples, such as [Participants | will record | pain and nausea post-operatively], ensure that the summary remains faithful to the article by capturing nuanced relationships. It does not just reiterate what the study does, but also how it goes about it, ensuring the reader understands the methodology.
    \item \textbf{Precision and Brevity}: The \method summary captures all the key points—right from the study's focus, the categorization of participants, to the post-operative documentation—without becoming verbose. It offers a condensed yet comprehensive view of the article, ensuring that readers can quickly grasp the core concepts without getting overwhelmed.
\end{itemize}

\section{Additional Evaluation}
\subsection{Performance on ClinicalTrial-Base}
In addition to the ClinicalTrial (Large) dataset, we also constructed a simpler version - ClinicalTrial-Base where we consider the article-summary pairs included in this dataset to be those with a BARTScore higher than $-2.0$. The statistics for this dataset are in Table \ref{tb:clinical_base_stat} shown as follows.
\begin{table}[!h]
\small
\centering
\resizebox{0.45\textwidth}{!}{
\begin{tabular}{lccccc}
\toprule
& \multicolumn{3}{c}{\textbf{\# Samples}} & \multicolumn{2}{c}{\textbf{\# Words}}   \\      
\textbf{Dataset} &Train &Valid &Test &Doc. &Sum. \\
\midrule
ClinicalTrial-Base &62,012 &7,752 &7,752 &277.7 &76.1 \\
\bottomrule
\end{tabular}
}
\caption{\textbf{Statistics of ClinicalTrial-Base.}}

\label{tb:clinical_base_stat}
\end{table}

\noindent Our evaluation results are shown in Table \ref{tb:perf_clinical_base} below.
\begin{table}[!h]
\small
\centering
\resizebox{0.5\textwidth}{!}{
\begin{tabular}{lcccc|cccc|cccc}
\toprule
 &\multicolumn{4}{c}{\textbf{ClinicalTrial-Base}}   \\  
\midrule
\textbf{Model}  & \textbf{R-1}   & \textbf{R-2}   & \textbf{R-L}   &$\Delta$  & \textbf{BS} & \textbf{BAS}    
\\    
\midrule
\textbf{Baselines} 
\\

$\textrm{T5}_{\textrm{Large}}$

&\textbf{53.9}      
&\textbf{41.7}      
&\textbf{47.2}      
&$-$2.0\% 

& 90.49
& -1.91
\\

\rowcolor{gray!20} ${\textrm{BART}_{\textrm{Large}}}$    

&51.8               
&38.6               
&43.6               
&$+$4.4\% 

& 89.61
& -1.99
\\

PEGASUS 

&51.8 
&40.7 
&44.8 
&$+$1.9\%

& 90.16
& -1.61
\\

$\textrm{GPT-3.5}_\textrm{zero-shot}$ 

&45.4 
&23.8 
&32.5 
&$+$37.6\%

& 89.00
& -2.44
\\

\midrule
\textbf{Our Method} \\
$\textrm{GPT-3.5}_\textrm{TriSum}$ 

&\textbf{54.1} 
&37.6 
&42.2 
&$+$4.5\%

& 90.84
& -1.52
\\

\texttt{TriSum-S}

&\textbf{53.6} 
&\textbf{42.2} 
&\textbf{46.6} 
&$-$1.8\%

& 90.67
& -1.66
\\

\texttt{TriSum-C}  

&50.3 
&37.2 
&42.8 
&$+$7.4\%    

& 89.25
& -2.14
\\

\texttt{TriSum-J}

&52.9 
&\textbf{41.8} 
&\textbf{45.2} 
& ---

& 90.81
& -1.64
\\
\bottomrule
\end{tabular}
}

\caption{\textbf{Performance comparison of ROUGE Scores and semantic similarity scores on ClinicalTrial-Base Dataset}.The top-3 results are \textbf{highlighted}. Our backbone model, ${\textrm{BART}_{\textrm{Large}}}$, is shadowed for reference.}

\label{tb:perf_clinical_base}
\end{table}

\subsection{More Baselines and Contrastive Learning Framework Adaptation}
\begin{table}[!h]
\small
\centering
\resizebox{0.48\textwidth}{!}{
\begin{tabular}{lccccc}
\toprule
\multicolumn{6}{c}{\textbf{CNN/DailyMail}}   \\  
\midrule
\textbf{Model} & \textbf{R-1}   & \textbf{R-2}   & \textbf{R-L} &\textbf{BS}  &\textbf{BAS}
\\    
\rowcolor{gray!20} $\text{BART}_{\text{Large}}$ 
& 44.0  & 21.1  & 40.6  &87.98  & -3.45    
\\
 $\text{BART}_{\text{12-6-SFT}}$ 
& 44.2  & 21.2  & 40.9  &88.04  & -3.47
\\
 $\text{PLATE}_{\text{BART 12-12},\lambda=2.0}$ 
& 44.9  & 22.0  & 41.4  &88.12  & -3.34
\\
 $\text{BRIO-Mul}_{\text{BART}}$ 
& 47.6  & 23.5  & 44.5  &88.74  & -3.22
\\
 $\text{LLAMA-2}_{\text{zero-shot}}$ 
& 36.4  & 14.2  & 30.4  &87.84  & -3.31
\\
\midrule
 $\texttt{TriSum}+\text{BRIO}_{\text{Mul}}$
&\textbf{48.0}   & \textbf{24.4} & \textbf{45.3}  &\textbf{89.38}   & \textbf{-3.07}
\\
 $\texttt{TriSum}_{\text{LLAMA-2}}$
& 45.5  & 22.7  & 42.0  &88.62  & -3.28
\\

\midrule
\multicolumn{6}{c}{\textbf{XSum}} \\
\midrule
\textbf{Model} & \textbf{R-1}   & \textbf{R-2}   & \textbf{R-L} &\textbf{BS}  &\textbf{BAS} 
\\
\rowcolor{gray!20} $\text{BART}_{\text{Large}}$  & 45.4  & 22.3  & 37.3  & 91.62 & -2.50
\\
 $\text{BART}_{\text{12-3-KD}}$ 
& 44.8  & 22.2  & 37.1  & 91.55 & -2.56
\\
 $\text{PLATE}_{\text{BART 12-12},\lambda=1.5}$ 
& 45.3  & 22.3  & 37.2  & 91.60 & -2.52
\\
 $\text{BRIO-Mul}_{\text{BART}}$ 
& 47.1  & 23.5  & 38.2  & 91.98  & -2.40
\\
 $\text{LLAMA-2}_{\text{zero-shot}}$ 
& 30.2  & 10.4  & 22.3  & 89.12  & -2.53
\\
\midrule
 $\texttt{TriSum}+\text{BRIO}_{\text{Mul}}$
& \textbf{48.2}  & \textbf{25.3}  & \textbf{39.9}  & \textbf{92.43}  & \textbf{-2.21}
\\
 $\texttt{TriSum}_{\text{LLAMA-2}}$
& 47.2  & 24.4  &39.3  &92.12  & -2.35
\\

\bottomrule
\end{tabular}
}
\caption{\textbf{Additional experiments.}}

\label{tb:perf_all_add}
\end{table}
\noindent In addition to Table \ref{tb:perf_all} and \ref{tb:perf_sem_sim}, we further tested baselines
$\text{BART}_{\text{12-3-KD}}$ \cite{shleifer2020pretrained} and 
$\text{PLATE}_{\text{BART 12-12},\lambda=1.5}$ \cite{zhang-etal-2022-attention}, a general contrastive learning-based framework
$\text{BRIO-Mul}_{\text{BART}}$ \cite{liu-etal-2022-brio}, and another leading LLM
LLAMA-2-70B \cite{touvron2023llama} on CNNDM and XSum datasets. We also tested \texttt{TriSum-J} (trained by GPT-3.5 rationale) futher trained with contrastive learning strategy from BRIO, denoted as ``$\texttt{TriSum}+\text{BRIO}_{\text{Mul}}$''. For a fair comparison, we use BART as backbone of BRIO for both datasets, while original paper of BRIO uses Pegasus for XSum. Moreover, we report
\texttt{TriSum} trained with the ``aspect-triple'' rationales generated by LLAMA-2-70B. We could not test with GPT-4's rationales due to the expensive API cost. Table \ref{tb:perf_all_add} presents our findings: (1) BRIO, as a general contrastive learning framework, can be adapted by \texttt{TriSum} and improve its performance, achieving SOTA results; (2) In a zero-shot scenario, LLAMA-2-70B outperforms GPT-3.5 on XSum; (3) \texttt{TriSum} shows comparable performance with both LLAMA-2 and GPT-3.5 rationales on the datasets.

\subsection{Fatualness Improvement with \texttt{TriSum}}
\begin{table}[!h]
\small
\centering
\resizebox{0.47\textwidth}{!}{
\begin{tabular}{lcccc}
\toprule
& BART  & \texttt{Trisum-J}   &$\text{GPT-3.5}_{\text{zero-shot}}$ &$\text{GPT-3.5}_{\text{TriSum}}$   \\  
\midrule
\textsc{FActScore}
& 88.1
& 92.9
& 85.3
& 93.7
\\  

\bottomrule
\end{tabular}
}
\caption{\textbf{Factual consistency evaluation on CNNDM test set.} Results will not affect the original paper's contributions.}

\label{tb:fact_consist}
\end{table}
\noindent We tested the \textbf{Factual Consistency} (FC) by \textsc{FActScore} \cite{min2023factscore} with their NP setting using Inst-LLAMA, and with the source text as the knowledge source. Table \ref{tb:fact_consist} shows that \texttt{TriSum} can substantially enhance FC, especially when using its rationale for GPT-3.5 prompting. This is because triples emphasizes the facts contained in the source text.
The result also indicates that, by systematically extracting the ``aspect-triple'' rationale, the model establishes a structured framework that constrains the generation process, minimizing the likelihood of generating content unsupported by the source text.

\end{document}